\pdfoutput=1
\documentclass[11pt]{article}
\usepackage{float}
\usepackage[preprint]{acl}

\usepackage{times}
\usepackage{latexsym}

\usepackage[T1]{fontenc}
\usepackage[utf8]{inputenc}

\usepackage{microtype}

\usepackage{inconsolata}

\usepackage{graphicx}
\usepackage{listings}
\usepackage{booktabs}   % \toprule \midrule \bottomrule
\usepackage{tabularx}   
\newcolumntype{Y}{>{\raggedright\arraybackslash}X}  %
\usepackage{inconsolata}
\usepackage{array}
\usepackage{pifont}
\usepackage{tabularx}
\usepackage{adjustbox}
\usepackage{multirow}
\usepackage{caption}
\usepackage{enumitem}
\usepackage{xspace}
\usepackage{tcolorbox}
\usepackage{booktabs,subcaption,amsfonts,dcolumn}
\usepackage{url}
\usepackage{amsmath,amsthm,amsfonts,amssymb,bm,stmaryrd, bbm}
\usepackage{color,xcolor,colortbl}
\usepackage{CJKutf8}
\usepackage{makecell}
\usepackage{booktabs}
\usepackage{graphicx}
\usepackage{listings}
\usepackage{stfloats}
\usepackage{float}
\usepackage{subcaption}
\usepackage{placeins}
\usepackage{dsfont}

\definecolor{warning}{HTML}{C80000}
\definecolor{mygreen}{rgb}{0.64, 0.76, 0.68}
\definecolor{myyellow}{rgb}{0.98, 0.94, 0.75}
\definecolor{mygreen}{rgb}{0.68, 0.85, 0.9}
\definecolor{myblue}{rgb}{0.82, 0.94, 0.75}
\definecolor{mypurple}{RGB}{224, 65, 245}
\definecolor{myorange}{RGB}{209, 136, 17}
\definecolor{Mycolor1}{HTML}{BAD8F2}
\definecolor{Mycolor2}{HTML}{DDEEFA}
\definecolor{Mycolor-green}{HTML}{CDE8CD}
\definecolor{Mycolor-red}{HTML}{FCEAEA} % light red

\usepackage[textsize=scriptsize]{todonotes}

\usepackage{afterpage}
\usepackage{enumitem}

\title{When Benchmarks Age: Temporal Misalignment through\\Large Language Model Factuality Evaluation}

\author{
Xunyi Jiang \quad Dingyi Chang \quad Julian McAuley \quad Xin Xu\thanks{Corresponding Author.}
\\ 
University of California, San Diego \\
\texttt{\{xuj003, dic006, jmcauley, xinxucs\}@ucsd.edu}
}

\usepackage{float}
\begin{document}
\maketitle
\begin{abstract}
The rapid evolution of large language models (LLMs) and the real world has outpaced the static nature of widely used evaluation benchmarks, raising concerns about their reliability for evaluating LLM factuality.
While substantial works continue to rely on the popular but old benchmarks, their temporal misalignment with real-world facts and modern LLMs, and their effects on LLM factuality evaluation remain underexplored.
Therefore, in this work, we present a systematic investigation of this issue by examining five popular factuality benchmarks and eight LLMs released across different years.
An up-to-date fact retrieval pipeline and three metrics are tailored to quantify benchmark aging and its impact on LLM factuality evaluation.
Experimental results and analysis illustrate that a considerable portion of samples in the widely used factuality benchmarks are outdated, leading to unreliable assessments of LLM factuality.
We hope our work can provide a testbed to assess the reliability of a benchmark for LLM factuality evaluation and inspire more research on the benchmark aging issue\footnote{Codes are available in \url{https://github.com/JiangXunyi/BenchAge}.}.
% \footnote{Code and data are available in the supplementary materials and will publicly release after reviewing.}

\end{abstract}

\section{Introduction}
New large language models have been released at an unprecedented pace recently~\citep{DBLP:journals/corr/abs-2303-18223, DBLP:journals/corr/abs-2402-06196}.
Accompanying this proliferation is the rise of numerous benchmarks that aim to compare diverse LLMs across a wide range of tasks~\citep{helm,DBLP:journals/tist/ChangWWWYZCYWWYZCYYX24,ma-etal-2025-pragmatics}.
Many benchmarks are static \citep{vuFreshLLMsRefreshingLarge2024}, meaning that the factual information they contain remains unchanged in response to real-world updates. 
For example, the answer to the question ``What is the most populated country in the world?'' is India\footnote{\url{https://en.wikipedia.org/wiki/List_of_countries_and_dependencies_by_population}} nowadays in 2025. 
However, the gold answer from SelfAware~\citep{yin-etal-2023-large} released in May 2023 is 
still China.
As a result, LLMs that provide up-to-date and factually correct answers may be unfairly penalized when evaluated against outdated benchmarks.
Despite this issue, numerous studies continue to rely on these static benchmarks to assess LLM factuality (\S \ref{sec:dataset_analysis}).
Although some works \citep{kasaiREALTIMEQAWhats,vuFreshLLMsRefreshingLarge2024} have introduced real-time updated benchmarks or methods to help LLMs obtain real-world information, the effects of using old benchmarks to evaluate present LLMs have not been systematically investigated (See Appendix \ref{app:related_work} for a detailed review of related works).

Motivated by this gap, we conduct a comprehensive empirical study on the temporal misalignment with five popular factuality benchmarks across different years and explore their implications for evaluating modern LLMs. 
Our study focuses on two key research questions: 
\begin{itemize}[nosep]
    \item \textbf{RQ1}: To what extent do widely used static benchmarks contain outdated factual answers compared to current real-world facts?
    \item \textbf{RQ2}: How does benchmark aging affect the factuality evaluation of modern LLMs?
\end{itemize}

To answer these questions, we focus on time-sensitive questions (\S \ref{sec:time-sensitive}).
We introduce a retrieval pipeline to obtain current real-world facts for temporal comparison (\S \ref{sec:retrieval}).
Three metrics, including \textbf{Dataset Drift Score, Evaluation Misleading Rate, and Temporal Alignment Gap},  are tailored to quantify temporal misalignment and measure the impacts of benchmark aging on LLM factuality evaluation (\S \ref{sec:temporal_comparison}).
% Specifically, we first extract time-sensitive questions from popular benchmarks and then search for the corresponding latest answers from the Internet. Finally, we compare and analyze the temporal misalignment between the gold benchmark answers, LLM responses, and searched answers.
% we introduce two quantitative measures that capture the effect of temporal drift: \textbf{Drift Score}, which measures the degree to which model outputs align with up-to-date search results rather than dataset answers, and \textbf{Factual Staleness} score, which measures the proportion of time-sensitive questions where model outputs and current facts are consistent with each other but differ from the benchmark answer, exposing instances of outdated knowledge. 
Extensive experiments are implemented across five commonly used benchmarks and eight LLMs released over different years. 
The results illustrate a considerable portion of samples in the old but popular benchmarks are outdated (RQ1).
The systematic analysis reveals that the temporal misalignment of the benchmarks with modern LLMs and the real-world facts will lead to unreliable assessments, raising concerns about the trustworthiness of LLM factuality evaluation (RQ2).
We hope that our work can provide a testbed to evaluate the benchmark reliability for factuality evaluation and inspire more considerations about benchmarking aging in future work.

\begin{figure*}[htbp]
\centering
\includegraphics[width=0.95\textwidth]{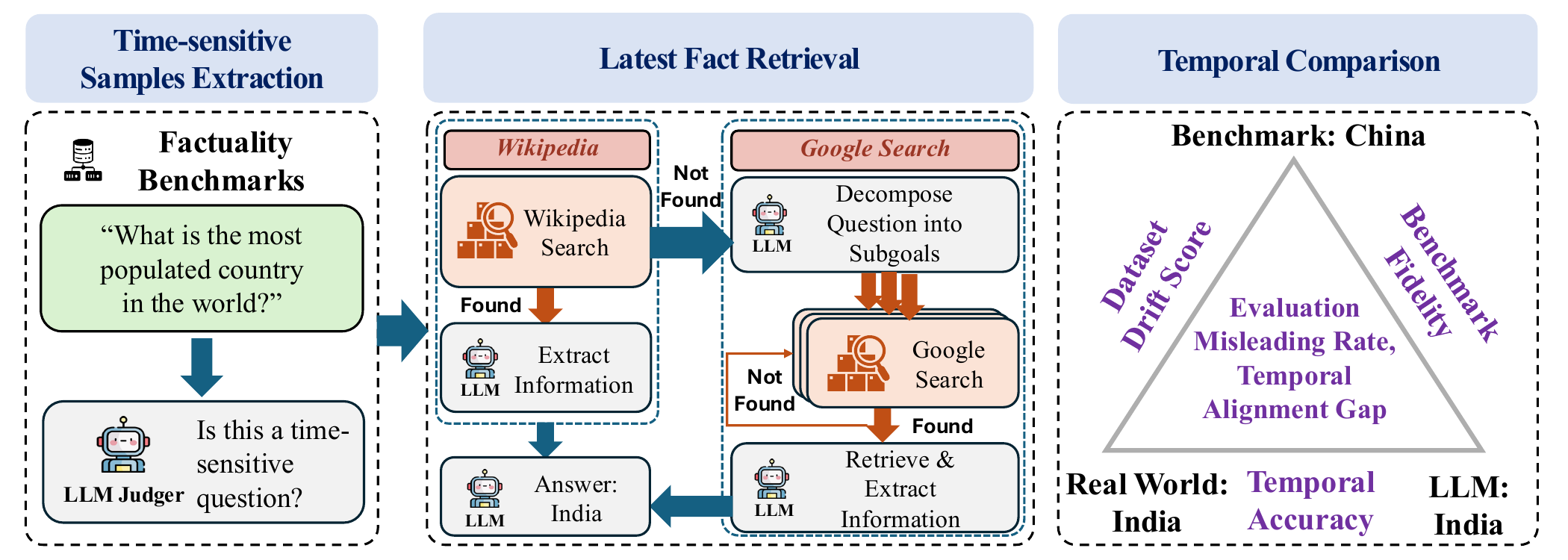}
\caption{Experimental setups. We first extract time-sensitive samples and then collect the corresponding real-world fact (with the latest fact retrieval pipeline), LLM response, and the gold label in the benchmark for each sample. Finally, we apply the proposed metrics to measure the temporal misalignment among them.}
\label{fig:pipeline}
\end{figure*}

\section{Experimental Setup}

% explore temporal misalignment, compare three
% 1. for real world, pipeline
% 2. comparison metrics: RQ1, RQ2
\label{sec:experimental}
To investigate temporal misalignment, we design a three-stage experiments to compare benchmark gold labels, LLM responses, and real-world facts.

\subsection{Time-sensitive Samples Extraction}
\label{sec:time-sensitive}
Focusing on temporal misalignment, we first extract \textit{time-sensitive samples}, which have verifiable factual answers that will change over time~\citep{weiMeasuringShortformFactuality2024b}, from the benchmarks.
The time-sensitive questions are identified for each benchmark by an LLM with human evaluations, achieving a 100\% recall and 90\% accuracy. The details are described in Appendix \ref{app:data_year_percentage} and \ref{app:detector-template}.
% For validation, manual evaluation by three domain experts on 150 questions yielded perfect recall (1.0) of time-sensitive questions.

\subsection{Latest Fact Retrieval}
\label{sec:retrieval}
To obtain the real-world facts, we retrieve the up-to-date answer from the Internet for each time-sensitive question.
% Large language models have shown strong performance in web search, summarization, and retrieval-augmented generation, making them effective for extracting information from the Internet \citep{chang-etal-2025-main}.
Our approach combines Wikipedia-focused retrieval and iterative web search, as depicted in Figure \ref{fig:pipeline}.
For each time-sensitive question, we first retrieve related information from Wikipedia, a widely regarded source of reliable factual information for popular topics and recent events~\citep{WikipediaAndAI}, using Brave Search\footnote{\url{https://brave.com/search/api/}}.
Secondly, GPT-4o-mini\footnote{\url{https://openai.com/index/gpt-4o-mini-advancing-cost-efficient-intelligence/}} is deployed to extract final answers from the retrieved information. 
If Wikipedia lacks suitable coverage, we use the Google Search API. 
Following ReAct and Chain-of-Action~\citep{yao2023reactsynergizingreasoningacting, panChainofActionFaithfulMultimodal2025}, we combine iterative reasoning with evidence retrieval. 
The system (1) decomposes questions into subgoals; (2) runs targeted searches for subgoals; (3) extracts key facts and temporal metadata; and (4) uses Qwen2.5-14B-Instruct~\citep{qwen2025qwen25technicalreport} to decide if further refinement and search are needed.
Detailed workflow is discussed in Appendix~\ref{human:web}. To assess the retrieval quality, each annotator with instruction in Figure~\ref{instct:web} manually reviewed 105 samples to determine whether the retrieved answers accurately reflect the searched evidence. The process achieves 89.52\% retrieval accuracy with moderate inter-annotator agreement ($\kappa=0.6$).

\subsection{Temporal Comparison}
\label{sec:temporal_comparison}
After obtaining the latest fact and the LLM response (Appendix \ref{appendix:experiment_groups}) for each time-sensitive question, we conduct a thorough analysis to explore temporal misalignment and its effects on LLM factuality evaluation.
We tailor the following metrics to help with analysis.
Specifically, given the query $x_i$ and the gold answer $y_i$ in each sample from the time-sensitive subset $\mathcal{D}_\text{ts}$ of a benchmark $\mathcal{D}$, the corresponding LLM response $\hat{y}_i$, and the real-world answer $y^*_i$ searched from the Internet, we compute two binary alignment scores:
% $s^{\text{gold}}_i(x_i, y_i, \hat{y}_i, y^*_i) \in \{0, 1\}$ is the agreement between $\hat{y}_i$ and $y_i$. 
$s^{\text{gold}}_i = \mathds{1}[\hat{y}_i=y_i]$ (the agreement between $y_i$ and $\hat{y}_i$ ), and $s^{\text{search}}_i = \mathds{1}[\hat{y}_i=y^*_i]$ (the agreement between $\hat{y}_i$ and $y^*_i$).

\begin{table*}[htbp]
\centering
\scalebox{0.825}{
\begin{tabular}{lccccc}
\toprule
\begin{tabular}[l]{@{}l@{}}\textbf{Dataset}\\ \small{Release Time}\end{tabular} & 
\begin{tabular}[c]{@{}c@{}}\textbf{TriviaQA}\\ \small{July 2017}\end{tabular} & 
\begin{tabular}[c]{@{}c@{}}\textbf{BoolQ}\\ \small{May 2019}\end{tabular} & 
\begin{tabular}[c]{@{}c@{}}\textbf{NaturalQuestion}\\ \small{July 2019}\end{tabular} & 
\begin{tabular}[c]{@{}c@{}}\textbf{TruthfulQA}\\ \small{May 2022}\end{tabular} & 
\begin{tabular}[c]{@{}c@{}}\textbf{SelfAware}\\ \small{July 2023}\end{tabular} \\
\midrule
% \textbf{Time-sensitive Ratio (\%)} & 2.22 & 13.76 & 10.19 & 19.58 & 11.15 \\
\textbf{Dataset Drift Score (\%)}       & 37.05 & 63.78 & 24.19 & 36.88 & 28.26 \\
\midrule
\midrule
\textbf{LLM} (Release Time) & \multicolumn{5}{c}{\textbf{Evaluation Misleading Rate (\%)}} \\
\midrule
Llama-2-7B-chat-hf (Jul 2023)        & 14.74 &  9.11 & 10.28 & 11.25 & 15.22 \\
Llama-3-8B-Instruct (Apr 2024)       & 11.16 &  8.22 & 10.28 &  8.13 & 19.57 \\
Llama-3.1-8B-Instruct (Jul 2024)     & 12.35 &  7.56 & 11.40 &  9.38 & 14.49 \\
Llama-3.2-3B-Instruct (Sep 2024)     &  9.16 &  8.67 &  9.52 & 10.63 & 10.51 \\
Ministral-8B-Instruct-2410 (Sep 2024) & 18.33 & 16.67 & 14.04 & 14.38 & 15.22 \\
GPT-4o-mini-2024-07-18 (Jul 2024)    & 19.92 & 17.11 & \colorbox{Mycolor-red}{24.06} & \colorbox{Mycolor-red}{23.13} & \colorbox{Mycolor-red}{22.10} \\
% Qwen2.5-1.5B-Instruct (Sep 2024)     &  9.96 & 10.67 &  5.51 &  8.13 & 13.41 \\
% Qwen2.5-3B-Instruct (Sep 2024)       & 10.36 & 13.56 & 10.03 & 16.25 & 18.48 \\
Qwen2.5-7B-Instruct (Sep 2024)       & 10.76 & 14.44 & 12.41 & 19.38 & 16.67 \\
% Qwen3-8B (April 2025) \\
Qwen2.5-14B-Instruct (Sep 2024)      & 13.55 & 16.00 & 16.04 & 16.88 & \colorbox{Mycolor-red}{22.46} \\
% \textbf{Average (EMR)}               & 12.97 & 11.78 & 11.95 & 13.40 & 16.18 \\
\bottomrule
\end{tabular}}
\caption{Dataset Drift Score (\%) of five widely-used LLM factuality benchmarks released along with time and Evaluation Misleading Rate (\%) of the modern LLMs (\colorbox{Mycolor-red}{>20\%}).}
\label{tab:main-table}
\end{table*}

\paragraph{RQ1} To quantify how the gold answers of time-sensitive samples in a benchmark have diverged from up-to-date real-world facts, we propose \textbf{Dataset Drift Score ($DDS$)}:
\begin{equation}
DDS = \frac{1}{|D_{ts}|} \sum_{i=1}^{|D_{ts}|} \mathds{1}[y_i \neq y^*_i],
\end{equation}

% EMR, DS (Appendix: BF, TA, CFR)
\paragraph{RQ2} We introduce two metrics to capture how benchmark aging affects LLM evaluation.
\textbf{Evaluation Misleading Rate ($EMR$)} reflects how often benchmark aging results in misleading evaluation results, which is the proportion of cases where an LLM gives an up-to-date answer, but is penalized by outdated benchmark labels:
\begin{equation}
\label{eq:2}
    EMR = \frac{1}{|D_{ts}|} \sum_{i=1}^{|D_{ts}|}\mathds{1}[\hat{y}_i = y_i^*  \land   \hat{y_i}  \neq y_i]
\end{equation} 
\textbf{Temporal Alignment Gap ($TAG$)} measures the discrepancy between LLM–world (Temporal Accuracy, $TA$) and LLM–benchmark (Benchmark Fidelity, $BF$):
\begin{align}
    TAG &= TA - BF \\
        &= \frac{1}{|D_{ts}|} \sum_{i=1}^{|D_{ts}|} s^{\text{search}} - \frac{1}{|D_{ts}|} \sum_{i=1}^{|D_{ts}|} s^{\text{gold}} \\
        &= \frac{1}{|D_{ts}|} \sum_{i=1}^{|D_{ts}|} (s^{\text{search}} - s^{\text{gold}})
\end{align}
A positive $TAG$ indicates that the LLM responses align more with the real-world facts than with the benchmark gold labels.

Based on these metrics, we investigate 8 diverse LLMs on commonly used LLM factuality benchmarks (Figure \ref{fig:citation}), including TriviaQA~\citep{joshiTriviaQALargeScale2017}, BoolQ~\citep{clarkBoolQExploringSurprising2019}, Natural Questions \citep{kwiatkowskiNaturalQuestionsBenchmark2019}, TruthfulQA \citep{linTruthfulQAMeasuringHow2022}, and SelfAware \citep{yinLargeLanguageModels2023}, which were released in different years.
The $DDS$ and $EMR$ are shown in Table \ref{tab:main-table}.
% To isolate the factors contributing to temporal drift,
% we designed three experimental groups.
% Since open-domain QA tasks can produce variable-length answers, we use the LLM-as-a-judge framework \citep{zheng2023judging} to standardize evaluation across models.
More details are shown in Appendix~\ref{appendix:experiment_groups} and \ref{app:llm-judge}.

\section{Experimental Results and Analysis}
\label{sec:results}

\subsection{RQ1: A Considerable Portion of the Benchmarks Are Outdated}

% LLM evaluation benchmarks are \textit{aging} due to the accumulation of time-sensitive questions and outdated gold answers.  
% We apply our time-sensitivity extraction module to five widely used QA benchmarks and find that

% The proportion of time-sensitive questions (specific numbers in Appendix \ref{app:data_year_percentage}) ranges from 2.22\% to 19.58\% in Table~\ref{tab:main-table}. 
% This indicates that while a few benchmarks remain relatively stable over time, most are substantially affected by temporal degradation.
 
$DDS$ values in Table~\ref{tab:main-table} show that at least 24.19\% (even up to 63.78\%) of time-sensitive samples are outdated as of July 19, 2025. 
Among all benchmarks, the relatively old benchmark, BoolQ, exhibits the highest DDS.
In contrast, newer benchmarks such as SelfAware show relatively less misalignment, reflecting shorter temporal distance from their release dates.
These results suggest that a considerable portion of the existing static and old benchmarks, though valid at release time, have become outdated over time.

\subsection{RQ2: Benchmark Aging Affects the Reliability of LLM Evaluation}
\label{sec:aging}

% \begin{figure*}[t]
%   \centering
%   % ---- Left: 58% ----
%   \begin{subfigure}[t]{0.5\textwidth}
%     \centering
%     \includegraphics[width=\linewidth,trim=2 10 8 6,clip]{plots/tag_grouped_bar_clean.pdf}
%     \caption{$TAG$ across the LLMs and benchmarks.}
%     \label{fig:tag}
%   \end{subfigure}\hfill
%   % ---- Right: 38% ----
%   \begin{subfigure}[t]{0.38\textwidth}
%     \centering
%     \raisebox{40pt}{\includegraphics[width=\linewidth]{plots/appendix_benchmark_citation_trend_polyfit_final_2025_10.pdf}}
%     \caption{Annual citation growth of the benchmark.}
%     \label{fig:citation}
%   \end{subfigure}

%   \caption{Comparison between temporal alignment metrics and dataset citation dynamics.}
%   \label{fig:fig2}
% \end{figure*}

% To measure how outdated benchmarks distort evaluation scores, we define the \textbf{Evaluation Misleading Rate} ($EMR$) in Equation~\ref{eq:2},  
% capturing the fraction of questions where model outputs align with real-world facts but diverge from the static benchmark labels. 
\paragraph{The outdated benchmarks can mislabel factually correct model responses.}
According to Table~\ref{tab:main-table}, more than half of the $EMR$ is larger than 10\%, indicating that a non-trivial fraction of LLM outputs are factually correct with respect to the real-world facts but judged as incorrect by stale benchmark labels. 
GPT-4o-mini and Qwen2.5-14B-Instruct exhibit relatively higher $EMR$ across all datasets than other LLMs, suggesting that newer LLMs are more vulnerable to evaluation bias, as they more frequently produce up-to-date answers that conflict with outdated references.
Overall, these results highlight that benchmark aging introduces systematic misalignment between factual model performance and reported evaluation scores.

\begin{figure}[t!]
    \centering
    \includegraphics[width=0.95\linewidth]{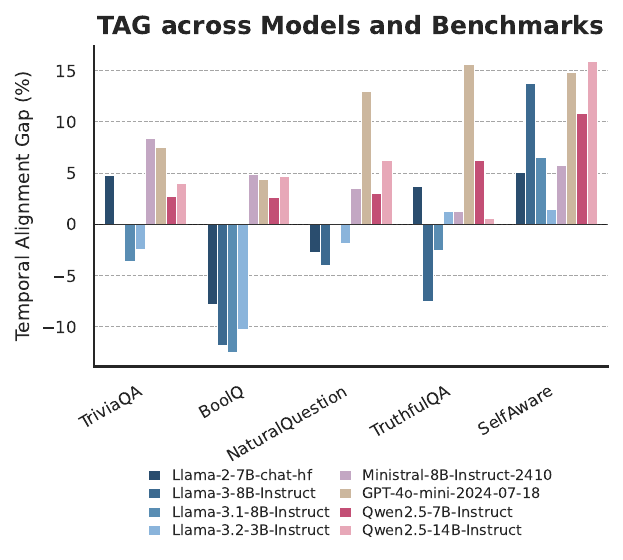}
    \caption{$TAG$ across the LLMs and benchmarks.}
    \label{fig:tag}
\end{figure}

\paragraph{The present LLMs are more aligned with real-world facts than with gold answers in the benchmarks.}
We further analyze the temporal consistency between LLM outputs, benchmark labels, and real-world facts through $TAG$ in Figure~\ref{fig:tag}, Table \ref{tab:temporal-accuracy}, and \ref{tab:benchmark-fidelity}.
Overall, 70\% of $TAG$ scores are positive, indicating that the LLMs mostly align more with up-to-date real-world facts than with outdated benchmark labels.
This pattern is observed consistently across the five benchmarks, especially for SelfAware, whose data are from relatively old datasets, such as SQuaD \citep{rajpurkar2016squad100000questionsmachine}, TriviaQA \citep{joshiTriviaQALargeScale2017}, and HotpotQA \citep{yang2018hotpotqa}.
Figure~\ref{fig:radar} shows Cohen's Kappa $\kappa$ \citep{cohenkappa} between each other among LLM responses, searched real-world information, and gold benchmark answers (Appendix \ref{app:cohen}).
The generally low $\kappa$ of LLM vs Gold ($<$0) and Gold vs Search (mostly $<$0.6) emphasizes the significant temporal misalignment between the gold labels in the benchmarks and the others.

\subsection{Dataset Analysis}
\label{sec:dataset_analysis}

\begin{figure}[t!]
    \centering
    \includegraphics[width=0.95\linewidth]{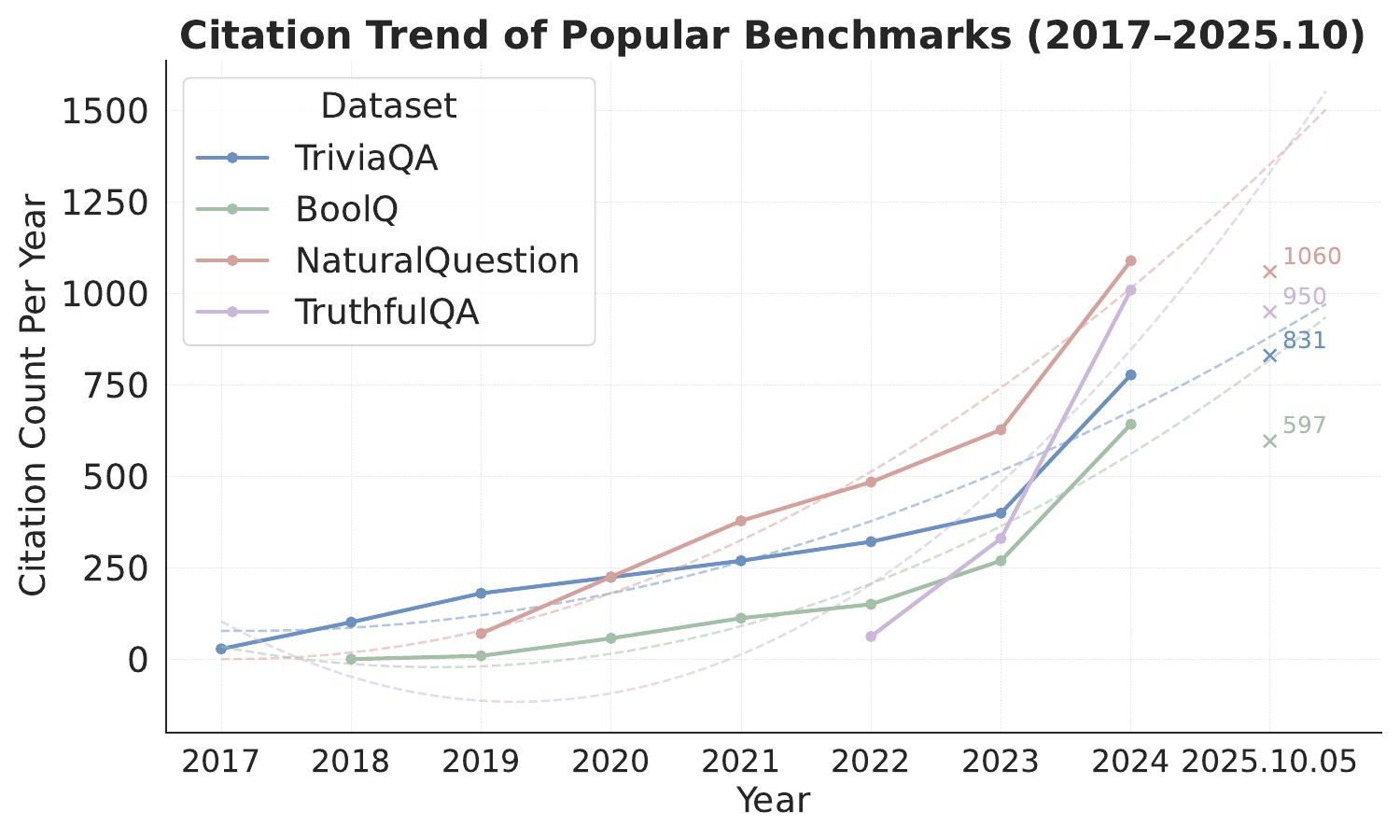}
    \caption{Annual citation growth of the benchmark.}
    \label{fig:citation}
\end{figure}

\paragraph{The usage of static benchmarks with outdated information is increasing.}
The release time in Table \ref{tab:main-table} suggests that the benchmarks we investigated are very old, and there is a significant time gap among the benchmarks, present LLMs, and real-world facts.
The upward trend in Google Scholar citations for these benchmarks is shown in Figure \ref{fig:citation}.
In the single year of 2024, the citations of Natural Questions and TruthfulQA surpassed 1,000, demonstrating their popularity for LLM evaluation.
These benchmarks have not been systematically updated to reflect evolving real-world facts. Nevertheless, they have been widely adopted in prior work and are likely to remain in use. This persistent reliance highlights the need for more attention to the unreliable use of the outdated benchmarks.

\paragraph{The outdated contexts amplify the temporal misalignment.} 
In open-book QA tasks, outdated information in the provided context can worsen factually temporal drift. 
\textsc{BoolQ} \citep{clarkBoolQExploringSurprising2019}, for instance, includes a supporting passage before a query.
As shown in Table~\ref{tab:boolq-drift}, models consistently exhibit more negative $TAG$ when performing passage-grounded inference. 
For example, Qwen2.5-7B-Instruct's $TAG$ drops from $2.67\%$ without the passage to $-12.22\%$ with it.
This indicates that the passages often encode outdated facts and anchor the model toward obsolete answers instead of correcting them since LLMs rely more on contexts instead of memorized knowledge \citep{li-etal-2023-large, zhou-etal-2023-context, DBLP:conf/iclr/Xie0CL024}, which suggests temporal degradation is not limited to open-ended generation but also affects passage-grounded evaluations.

% \subsection{Discussion}
% % The answers above for RQ1 and RQ2 indicate a significant temporal misalignment issue when researchers use static benchmarks to evaluate present LLMs.
% % We hope our discussion can highlight the importance of this issue and give inspiration for LLM evaluation in future work.

% Our results show that benchmark aging causes systematic bias in factuality evaluation.  
% As datasets drift away from real-world facts, newer models are sometimes penalized for giving correct answers.  
% This misalignment, measured by $DDS$, $EMR$, and $TAG$, appears in both closed-book and passage-grounded QA.  
% Similar trends have been reported in dynamic benchmarks such as \textsc{FRESHQA} \citep{vuFreshLLMsRefreshingLarge2024} and \textsc{RealTimeQA} \citep{kasaiREALTIMEQAWhats}, highlighting the necessity of continual temporal auditing.
% We suggest two practical remedies:  
% (1) \textbf{Temporally-aware benchmark design}: Reduce temporal sensitivity during dataset creation by adding explicit temporal scopes (e.g., “as of 2023”) or by including temporal metadata that allows answers to be re-validated when used in future evaluations. 
% (2) \textbf{Automatic benchmark maintenance}: periodically audit and refresh gold answers using reliable, updated knowledge sources (e.g, Wikipedia), enabling benchmarks to stay aligned with real-world facts. 
% These steps can greatly reduce the bias caused by outdated dataset labels and improve the reliability of factual evaluation as the dataset ages. 

\section{Conclusion}
In this work, we conduct a comprehensive empirical study and provide a testbed to investigate the temporal misalignment of the existing static LLM benchmarks with present LLMs and the real world, and its impacts on LLM factuality evaluation.
% Extensive experiments are conducted with five popular benchmarks and eight LLMs released across years.
The results and analysis reveal that a considerable portion of samples in the widely used factuality benchmarks are outdated.
Reliance on these aging benchmarks will lead to unreliable LLM factuality evaluation.
We hope this work can suggest future research to consider temporal misalignment in the benchmark design and LLM factuality evaluation.

\newpage
\section*{Limitations}
Despite our comprehensive analysis, some limitations may still remain in this study.
Our experiments are limited to a representative subset of available modern LLMs due to computational resources and budget constraints. We primarily evaluated models from a few open-source families and did not include larger LLMs, reasoning-based LLMs, and more commercial models, which may exhibit different temporal knowledge behaviors. Future work should expand the scope to validate and generalize our findings.
We only conducted our evaluation on five widely used factuality benchmarks. Although these benchmarks cover a range of tasks and domains, they may not fully represent all areas where temporal misalignment impacts model evaluation. More research on additional datasets will be considered to better understand the generality of temporal drift effects in future work.
% Our pipeline relies on LLMs as assistants in search and retrieval tasks, and as judges to assess alignment between model outputs and current facts. Although humans verify that LLM-based assessments generally align with human judgments, these automated evaluations may still be imperfect.

% more models, reasoning models, conmercial models
% five datasets recent ones and 
% llm-as-a-judge
\section*{Ethical Considerations}
% release the code and datasets, the materials will be public, copyright, search results are all from public resources
We recognize the importance of transparency and reproducibility in research.
We publicly release all code, data, and evaluation materials in this study after the review period. Our searched data is sourced exclusively from publicly available resources, with full citations provided to ensure proper attribution and traceability. By making these materials accessible, we aim to support future research while respecting copyright and intellectual property rights.

\section*{Acknowledgment}
The authors thank Aakash Agrawal for his metric designs and review of the paper.
This work is partially supported by NSF IIS-2432486.

\bibliography{main}

% \newpage
\appendix

\section{Related Works}
\label{app:related_work}
The evaluation of large language models (LLMs) relies heavily on standardized benchmarks, which provide a common ground for comparing model performance across tasks and over time. Benchmarks such as Natural Questions \citep{kwiatkowskiNaturalQuestionsBenchmark2019} and TriviaQA \citep{joshiTriviaQALargeScale2017} are widely used for evaluating LLMs, as they enable standardized comparisons, offer an objective measure of model performance, and help track advances in model capabilities. However, most existing benchmarks are static, capturing a snapshot of knowledge at a particular point in time, and thus may not reflect the evolving nature of real-world information.

As factual knowledge evolves, static benchmarks can quickly become outdated, leading to differences between what a benchmark evaluates and the current state of the world. This temporal misalignment is highlighted in recent studies showing that LLMs may provide correct answers according to up-to-date information yet be penalized by benchmarks anchored to outdated facts \citep{kasaiREALTIMEQAWhats, vuFreshLLMsRefreshingLarge2024}. To address this, several works propose dynamic or regularly updated benchmarks, such as RealTimeQA \citep{kasaiREALTIMEQAWhats} and FreshQA \citep{vuFreshLLMsRefreshingLarge2024}, which are designed to evaluate models on time-sensitive questions and recent events. In a complementary direction, \citet{luu-etal-2022-time} investigate temporal drift from a model-centric perspective, analyzing how model performance changes when trained and tested on temporally shifted datasets. Recently, WiNELL \citep{winell} tackles temporal misalignment from the data side by maintaining up-to-date content. These efforts emphasize the importance of incorporating temporal dynamics into benchmark design to ensure accurate and meaningful LLM evaluations.

Beyond temporal misalignment, recent research has revealed fundamental flaws in benchmark construction and evaluation methodologies that compromise the reliability of performance assessments. Systematic audits of popular reasoning benchmarks, including SocialIQa \citep{DBLP:journals/corr/abs-1904-09728}, FauxPas-EAI \citep{shapira-etal-2023-well}, and ToMi \citep{le-etal-2019-revisiting}, have uncovered issues across many dimensions \citep{garbageinreasoningout}. These studies show that a substantial portion of LLM errors are attributable not to the model, but to flaws in benchmark design, including duplicated items, ambiguous wording, implausible answers, and scoring procedures that prioritize output format over reasoning process. These findings challenge the validity of current benchmark-based claims about reasoning in LLMs and highlight the need for evaluation protocols that assess intended capabilities rather than unfairly penalize models due to benchmark design flaws or outdated factual content.

% \subsection{Automated Knowledge Base Updating with LLMs}
% Recent advances in LLM-based agents have enabled new approaches to automated knowledge base maintenance, particularly for continuously evolving repositories like Wikipedia. WINELL (Wikipedia Never-Ending Language Learning) represents a significant advancement in this domain, introducing an agentic framework for autonomously updating Wikipedia articles through online information aggregation \citep{winell}. Their system addresses the fundamental challenge of maintaining up-to-date content in Wikipedia, which relies heavily on manual human editors and often suffers from latency in incorporating new information. The system employs LLM-based agents that use iterative web searching and information aggregation to identify relevant updates, addressing the temporal misalignment problem in their knowledge source by continuously monitoring online sources for new information. WINELL's approach demonstrates how automated systems can reduce the latency between knowledge bases and updated information, providing a complementary perspective to dynamic benchmark evaluation methods that seek to address similar temporal challenges in LLM assessment.

% \clearpage
\section{Dataset Information}
\subsection{Dataset Creation Year and Time-sensitive Percentage}
\label{app:data_year_percentage}
Table~\ref{tab:ts-datasets} summarizes the statistics of QA datasets used in this paper, including release year, total QA pairs, percentage of time-sensitive (TS) questions, and answer types. The time-sensitive questions extraction details are shown in Appendix~\ref{app:detector-template}.  Aside from TriviaQA, which only contains 2.22\% time-sensitive questions, other datasets contains more than 10\% of time-sensitive data. This shows that a non-negligible proportion of time-sensitive data exists in these popular benchmarks.

\begin{table}[t]
\centering
\small
\setlength{\tabcolsep}{4pt}

\begin{tabular}{lccc}
\toprule
\textbf{Property} & TriviaQA & BoolQ & NaturalQuestion \\
\midrule
\textbf{Year} & 2017 & 2019 & 2019 \\
\textbf{\# QA Pairs} & 11,313 & 3,270 & 7,830 \\
\textbf{TS \%} & 2.22 & 13.76 & 10.19 \\
\textbf{Type} & Open QA & Multi-choice & Open QA \\
\bottomrule
\end{tabular}

\vspace{0.5em}
\begin{tabular}{lcc}
\toprule
\textbf{Property} & TruthfulQA & SelfAware \\
\midrule
\textbf{Year} & 2022 & 2023 \\
\textbf{\# QA Pairs} & 817 & 2,475 \\
\textbf{TS \%} & 19.58 & 11.15 \\
\textbf{Type} & Open QA & Open QA \\
\bottomrule
\end{tabular}
\caption{Overview of QA datasets with time-sensitive (TS) questions.}
\label{tab:ts-datasets}
\end{table}

\subsection{Google Scholar Citation Trend}
To estimate the influence of each benchmark, we measure its citation trend using Google Scholar \footnote{\url{https://scholar.google.com/}}. The citation data we collect is as of Oct 3, 2025. Specifically, we record the number of "cited by" results with year-specific filters from 2017 to 2025. In order to measure the future prediction trend, we use a polynomial to predict the citations at the end of 2025. In 2024 single year, the summation of citations for these 4 datasets is 3,521, revealing consistent scholarly interest in natural language processing and factuality question answering task.
% \begin{figure}[htbp]
%     \centering
%     \includegraphics[width=\linewidth]{plots/appendix_benchmark_citation_trend_polyfit_final_2025_10.pdf}
%     \caption{Trends of yearly Google Scholar citation for popular LLMs Benchmarks.}
%     \label{appendix:citation-trend}
% \end{figure}

% \FloatBarrier
% \begin{figure}[H]
%     \centering
%     \includegraphics[width=\linewidth]{/plots/time_sensitive_label_pipeline.pdf}
%     \caption{Process for determining the time-sensitivity of dataset questions.}
%     \label{time_sensitive_pipeline}
% \end{figure}
% \FloatBarrier

% \noindent\makebox[\textwidth]{%
%     \includegraphics[width=0.7\paperwidth]{/plots/time_sensitive_label_pipeline_v.pdf}%
% }

% \captionof{figure}{Process for determining the time-sensitivity of dataset questions.}
% \label{time_sensitive_pipeline}
% \clearpage

\section{Experiment Details}

\subsection{Experiment Setups}
\label{appendix:experiment_groups}
\begin{enumerate}
    \item \textbf{Computation environment}: All experiments were conducted on an Ubuntu~22.04 workstation equipped with four~NVIDIA~RTX~A6000 GPUs (48\,GB each), running CUDA~12.6 and PyTorch~2.7.0{+}cu126. 
    The software stack includes Python~3.10.18, Transformers~4.53.1, and OpenCompass~0.5.0 for evaluation orchestration, together with vLLM~0.9.2 for optimized inference. 
    % Experiments were managed under a dedicated Conda environment and verified reproducibility across random seeds and GPU devices. 

    \item \textbf{Model architectural analysis}: Models from different architectures released in a similar timeframe (June-October 2024): Qwen2.5-7B-Instruct \footnote{\url{https://huggingface.co/Qwen/Qwen2.5-7B-Instruct}}, Ministral-8B-Instruct-2410 \footnote{\url{https://huggingface.co/mistralai/Ministral-8B-Instruct-2410}}, Llama-3.1-8B-Instruct \footnote{\url{https://huggingface.co/meta-llama/Llama-3.1-8B-Instruct}}, and GPT-4o-mini \footnote{\url{https://openai.com/index/gpt-4o-mini-advancing-cost-efficient-intelligence/}} . Controlling for release date isolates the effects of architectural differences on temporal knowledge retention.
    % \citep{touvron2023llama2openfoundation, grattafiori2024mistralmodels, meta2024llama32}. \citep{qwen2025qwen25technicalreport, mistral2024ministraux, grattafiori2024mistralmodels}
    
    % \item \textbf{Training data cutoff analysis}: Llama-family models with comparable parameter sizes (7B-8B) but different release dates, from Llama-2-7B \footnote{\url{https://www.llama.com/llama2/}}(January 2023) to Llama-3.2-3B-Instruct (September 2024). This group examines the impact of data recency.
    
    \item \textbf{Model scale analysis}: Qwen2.5 models of varying sizes (1.5B, 3B, 7B, and 14B) released simultaneously in September 2024 \citep{qwen2025qwen25technicalreport}, isolating the effect of model scale. 
\end{enumerate}
\subsection{Time-sensitive Samples Extraction}
\label{app:detector-template}
\begin{figure}[h]
    \centering
    \includegraphics[width=\linewidth]{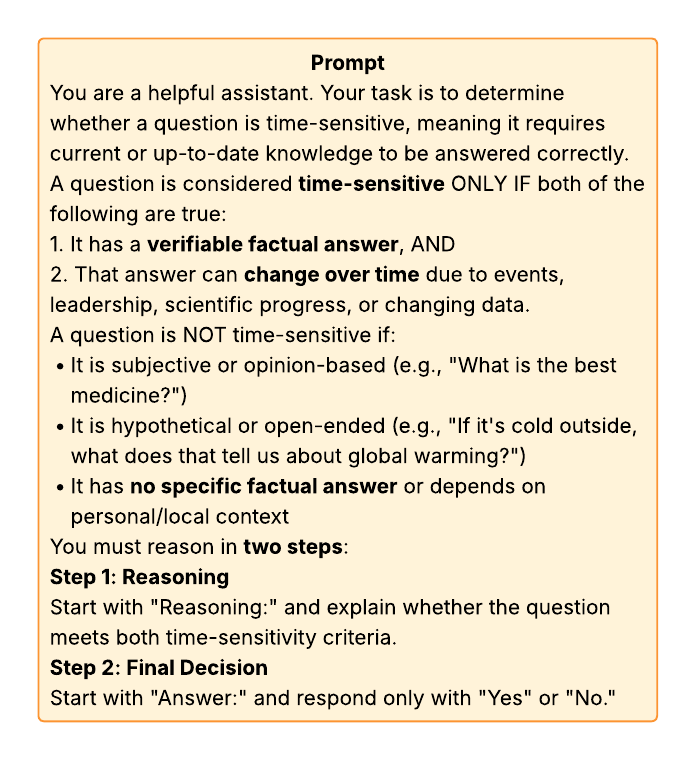}
    \caption{Prompt for Determining the Time-Sensitivity of Dataset Questions.}
    \label{time_sensitive_pipeline}
\end{figure}

We use Qwen-2.5-14B-Instruct to extract time-sensitive samples from existing QA benchmarks. Specifically, we serve Qwen-2.5-14B-Instruct using the vLLM framework\footnote{\url{https://github.com/vllm-project/vllm}} for efficient inference. To reduce the randomness in LLM responses when identifying time-sensitive questions, we apply the prompt shown in Figure~\ref{time_sensitive_pipeline} three times independently and determine the final label by majority voting. We then conduct a grid search over the model’s temperature and the number of voting rounds. Our results show that using three votes with a temperature of 1.0 yields the highest accuracy 90\% while maintaining 100\% recall for time-sensitive questions.
\begin{figure}[h]
    \centering
    \includegraphics[width=\linewidth]{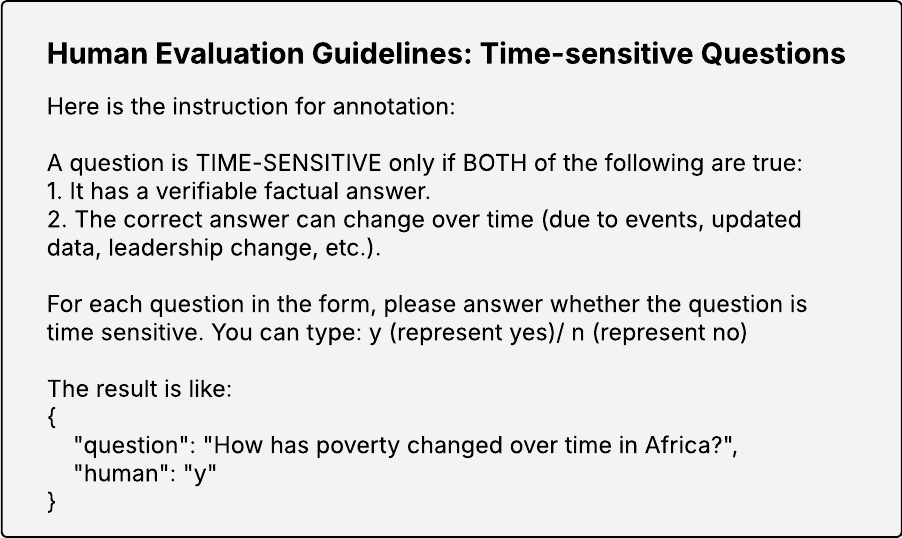}
    \caption{Instructions for Human Evaluation of Time-sensitive Questions}
    \label{instct:ts}
\end{figure}

To further validate the labeling quality, we conduct a human evaluation. Given the instruction in Figure~\ref{instct:ts}, three domain experts manually annotate 150 questions, with results presented in Table~\ref{tab:ts-eval}. All annotations were performed by graduate-level NLP researchers from our institution. These annotators are fluent in English and have prior experience evaluating QA datasets. Since the annotation task involved only publicly available benchmark data, no new human subject data was collected. Importantly, no crowd-sourcing platforms were involved; instead, the annotators participated voluntarily without any financial compensation. As a result, issues of participant recruitment, payment fairness, or data consent do not apply. Nonetheless, the annotation process and data usage were reviewed internally to ensure ethical compliance.

\begin{table}[h]
\centering
\scalebox{0.8}{
\begin{tabular}{lccccc}
\toprule
\textbf{Metric} & Recall & F1 Score & Accuracy & Cohen's Kappa \\
\midrule
\textbf{Score} & 1.000 & 0.909 & 0.9 & 0.83375 \\
\bottomrule
\end{tabular}}
\caption{Human evaluation of time-sensitive question detection.}
\label{tab:ts-eval}
\end{table}

\subsection{Web Search Pipeline}
\begin{figure}[htbp]
    \centering
    \includegraphics[width=0.8\linewidth]{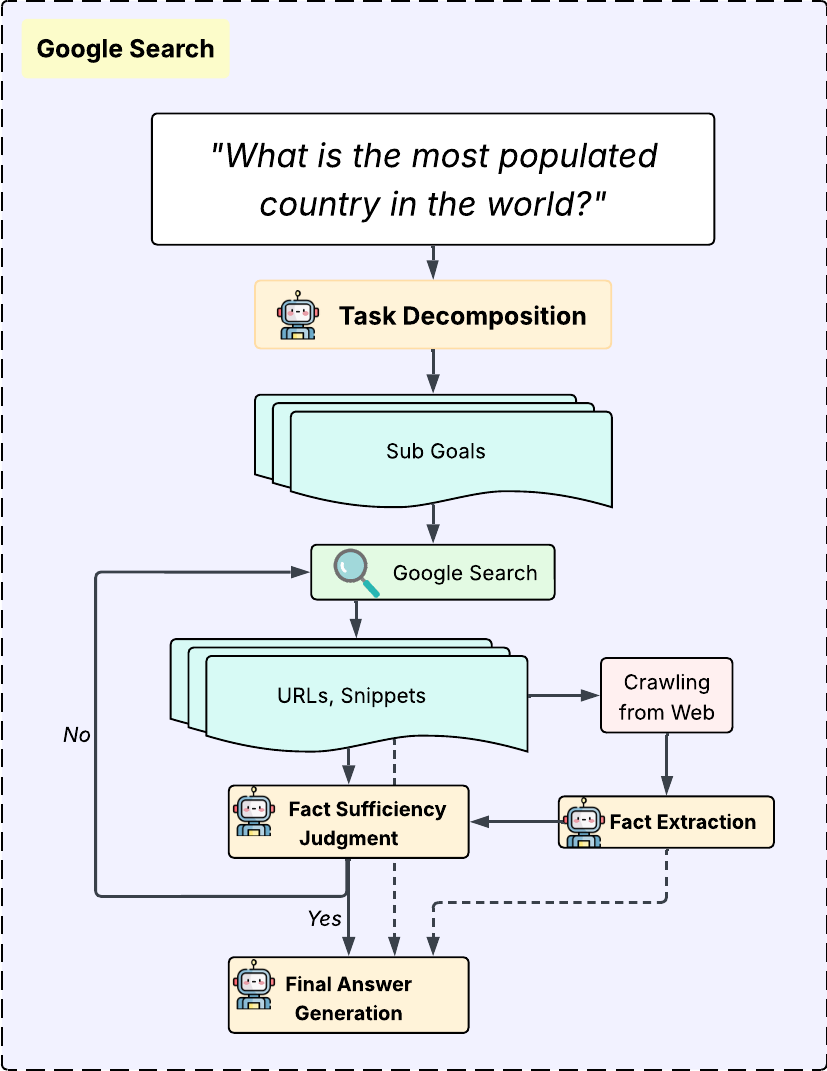}
    \caption{Workflow of Google Search and Fact Retrieval.}
    \label{fig:google_search}
\end{figure}

All web search results were collected during a fixed time window from July 18 to July 19, 2025, ensuring consistency and temporal alignment across all queries. We utilize both the Google Search API and Brave Search (which includes access to Wikipedia content) to retrieve supporting evidence from the open web. To ensure robustness, our system is designed to tolerate transient network errors and incomplete results. In practice, we implement a retry mechanism: for Brave search, we retry up to three times in the event of failure. For Google search, as shown in Figure~\ref{fig:google_search}, we repeat the search process adaptively until either sufficient information is found (as judged by the LLM) or a hard limit of 15 search attempts is reached. These search engines are chosen for their broad coverage, freshness, and reliability—especially valuable for capturing real-world updates that static benchmarks fail to reflect.

This design reflects the reality that many benchmark answers cannot be verified from a single source like Wikipedia. Our logs reveal that only 22.3\% of our selected questions were retrieved using Wikipedia as the source. The remainder required external evidence. To address this, our pipeline combines search engines, evidence consolidation, and LLM-based filtering, ensuring higher precision. We report the sources and methods used for transparency, and emphasize that our pipeline is intended for analysis of benchmark staleness, not as a replacement for routine benchmark updating. Notably, FreshQA also relies on web search to retrieve the latest answers for their benchmark questions, underscoring that web-based retrieval is a practical and accepted strategy for keeping benchmarks aligned with real-world facts.

\vspace{0.5em}

\begin{figure}[htbp]
    \centering
    \includegraphics[width=\linewidth]{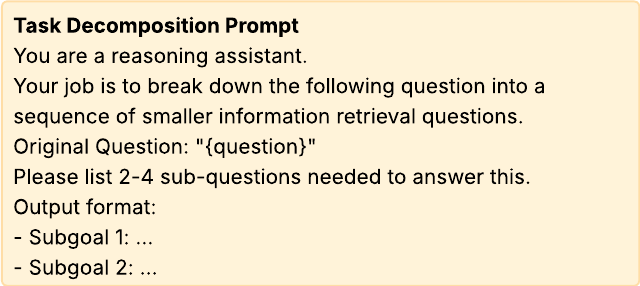}
    \caption{Task Decomposition Prompt}
    \label{fig:prompt-task}
\end{figure}

\begin{figure}[htbp]
    \centering
    \includegraphics[width=\linewidth]{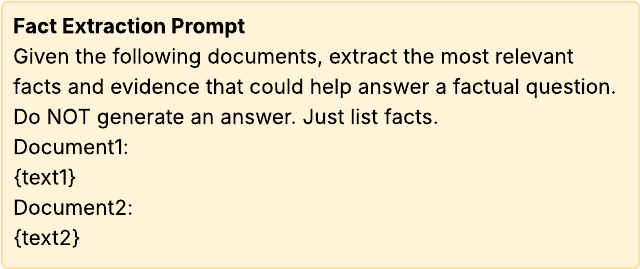}
    \caption{Fact Extraction Prompt}
    \label{fig:prompt-fact}
\end{figure}

\begin{figure}[htbp]
    \centering
    \includegraphics[width=\linewidth]{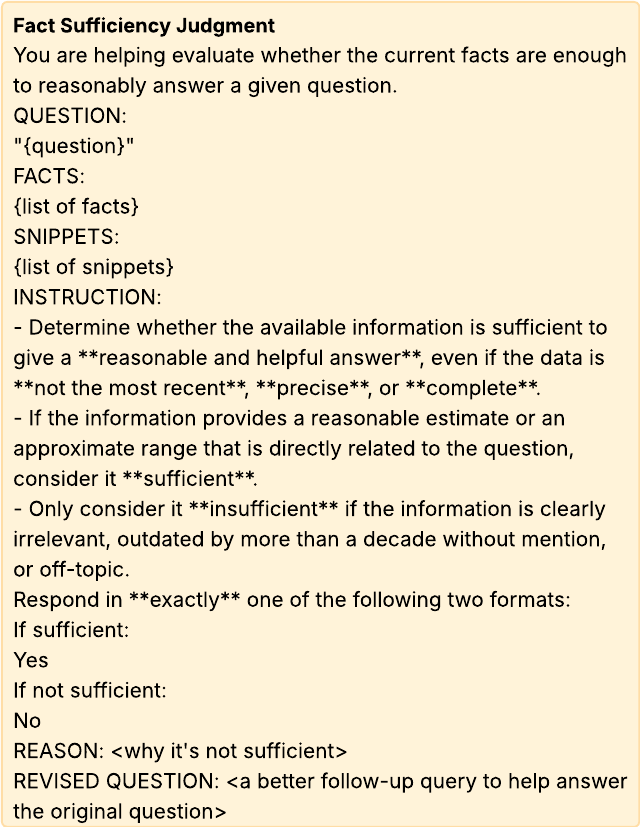}
    \caption{Fact Sufficiency Judgment Prompt}
    \label{fig:prompt-suff}
\end{figure}

\begin{figure}[htbp]
    \centering
    \includegraphics[width=\linewidth]{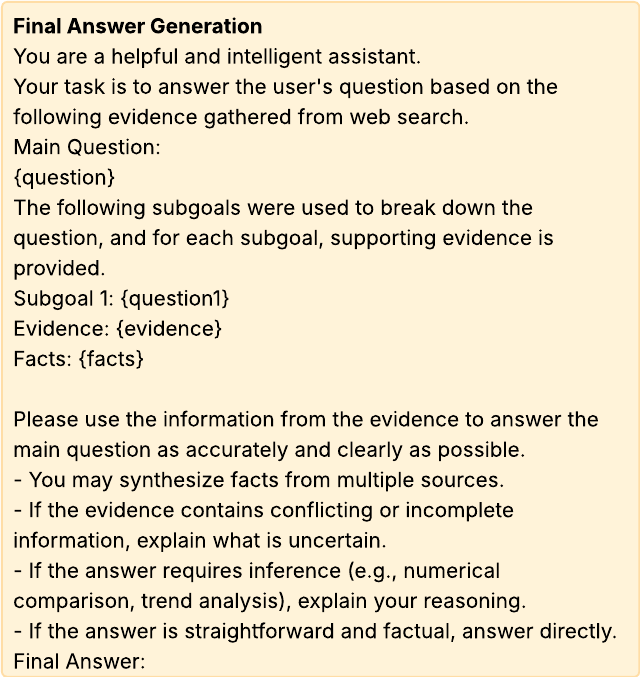}
    \caption{Final Answer Generation Prompt}
    \label{fig:prompt-final}
\end{figure}

To support the retrieval and reasoning process, we design a set of LLM prompts tailored to each stage of the pipeline. These prompts guide the model through subgoal planning, evidence extraction, fact sufficiency evaluation, and final answer generation. Visualizations of the four prompt templates are shown in Figures~\ref{fig:prompt-task}--\ref{fig:prompt-final}.

\begin{figure}
    \centering
    \includegraphics[width=\linewidth]{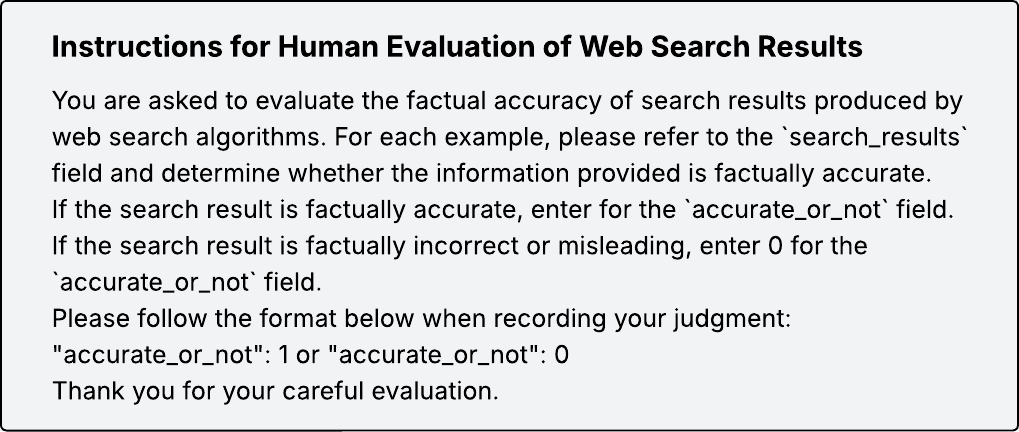}
    \caption{Human Evaluation Instruction for Web Search Results}
    \label{instct:web}
\end{figure}

To assess the quality of web search output, we adopt two complementary methods:
\label{human:web}
\paragraph{Human Evaluation:} We randomly sample 105 questions from the dataset and ask three domain experts to manually assess whether the web search outputs provide factually correct answers. The overall accuracy reaches 89.52\%, indicating a high degree of factual consistency. Furthermore, inter-annotator agreement, measured using Cohen’s Kappa, is 0.58, which reflects moderate agreement. The detailed annotation instruction is provided in Figure~\ref{instct:web}.

\paragraph{Cohen’s Kappa Analysis:} To further evaluate the alignment between web search results and the dataset's gold answers, we calculate Cohen’s Kappa scores across all time-sensitive questions. As illustrated in Figure~\ref{fig:radar}, the green polygon representing this agreement lies between 0.5 and 0.8, suggesting a relatively strong consistency between search-derived answers and dataset labels. This level of agreement is expected, as only a small portion of questions involve fast-changing knowledge. Therefore, we infer that the retrieved web search results are generally reliable and can serve as a valid approximation of current factual information.

\subsection{LLM-as-a-judge Prompt}
\label{app:llm-judge}
\begin{figure}[h]
    \centering
    \includegraphics[width=\linewidth]{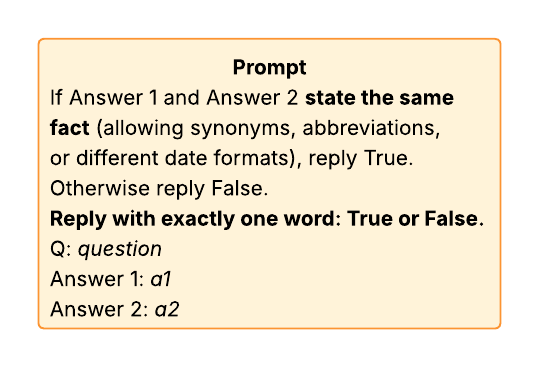}
    \caption{Prompt for determining the time-sensitivity of dataset questions.}
    \label{fig:llm_judge_prompt}
\end{figure}

% To measure the agreement between model responses, benchmark answers, and real-world information, we use Cohen's Kappa coefficient as an inter-rater reliability metric.
% Using the LLM-as-a-judge framework, we convert each source's response into categorical alignment scores, treating them as independent evaluators. 
% This yields a robust, style-invariant measure of factual alignment across temporally diverse knowledge sources. We leverage a simple and interpretable prompt to let LLMs determine factual consistency between answers.
To evaluate how well the model responses, benchmark answers, and real-world information agree with each other, we use Cohen’s Kappa coefficient. This metric measures how consistently different sources align in their answers. We treat each source—the model, the benchmark, and the web search result—as an independent evaluator. Using the LLM-as-a-judge setup, we apply a clear and interpretable prompt in Figure~\ref{fig:llm_judge_prompt} that asks the LLM to judge whether two answers express the same factual content. This process allows us to convert the answers into simple agreement scores, giving us a reliable and style-independent way to compare factual consistency across sources that may reflect knowledge from different points in time.

\begin{figure}
    \centering
    \includegraphics[width=\linewidth]{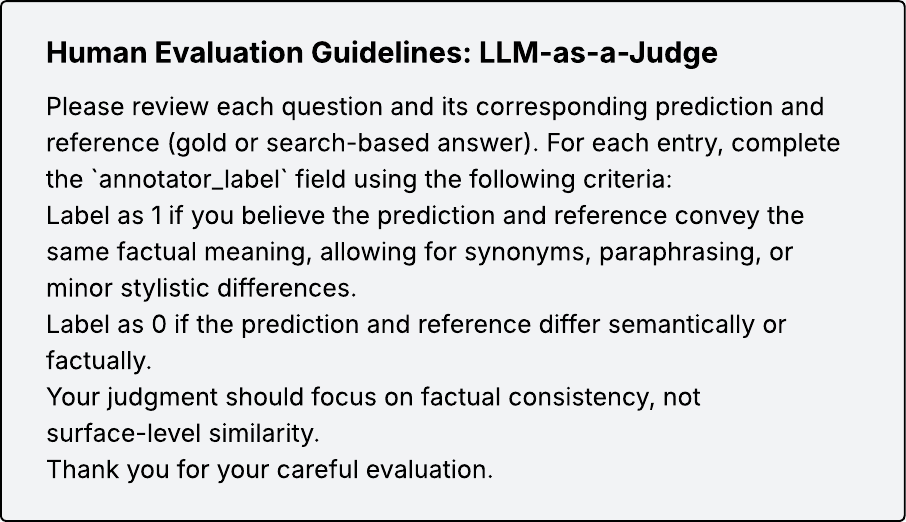}
    \caption{Human Evaluation Instructions for LLM-as-a-judge}
    \label{instct:judge}
\end{figure}

Similar to time-sensitive classification and websearch, we perform human evaluation of LLM-as-a-judge with the instructions in Figure~\ref{instct:judge}. Outputs yields the following agreement: the accuracy is 97\% and the average Cohen's Kappa between three evaluators is 0.72.

\section{Experiment Results}
\subsection{Cohen's Kappa Score}
\label{app:cohen}
To systematically evaluate the agreement between different information sources, we compute the Cohen’s Kappa coefficient, a standard inter-rater reliability metric in statistics and NLP. Formally, Cohen’s Kappa is defined as $\kappa = \frac{p_o - p_e}{1 - p_e}$, where $p_o$ is the observed agreement and $p_e$ the expected agreement by chance. Unlike raw accuracy, Cohen's Kappa adjusts for chance-level agreement and thus provides a more robust and interpretable measure of consistency across different answer sources.

\begin{figure*}[htbp]
    \centering
    \includegraphics[width=1\linewidth]{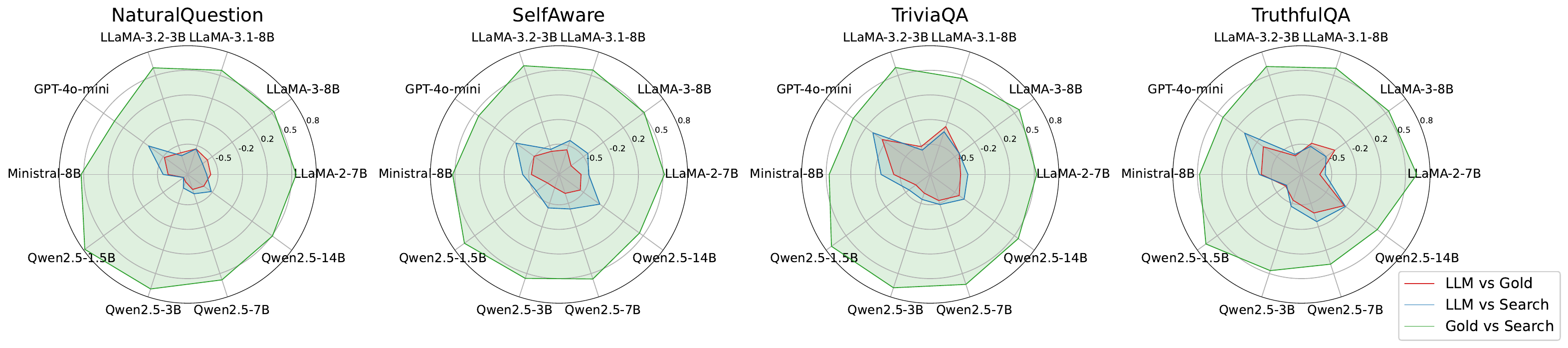}
    \caption{Cohen's Kappa Score between each other among LLMs' responses, searched real-world information, and gold benchmark answers from datasets and LLMs.}
    \label{fig:radar}
\end{figure*}

Figure~\ref{fig:radar} presents a radar plot of pairwise Cohen’s Kappa scores among model outputs, web search results, and benchmark gold labels, computed across four datasets and ten representative LLMs. The radar shape reveals several insights. First, the agreement between web search and gold answers is generally high, indicating that our retrieval pipeline reliably captures accurate, up-to-date information. Second, the agreement between LLMs and the benchmark is lower, suggesting possible misalignment due to temporal drift or limitations in training data coverage. Finally, the agreement between LLMs and web search tends to be more variable, highlighting the inconsistent ability of models to match real-world facts in time-sensitive contexts.

Overall, this analysis illustrates the discrepancy between static benchmarks, dynamic web content, and model outputs. It motivates the need for time-aware evaluation and fact-checking frameworks that consider real-world knowledge freshness.

\subsection{Temporal Accuracy and Benchmark Fidelity}
We calculate $TAG$ from Temporal Accuracy and Benchmark Fidelity. $TA$ is shown in Table~\ref{tab:temporal-accuracy}. GPT-4o-mini-2024-07-18 still performs best overall datasets. This represents that GPT-4o-mini-2024-07-18 contains most up-to-date real world information. As the only close-source model, this observation highlights that currently the commercial model update more frequently.
\begin{table*}[htbp]
\centering
\scalebox{0.9}{
\begin{tabular}{lccccc}
\toprule
\textbf{Model / Dataset (Release Time)} &
\begin{tabular}[c]{@{}c@{}}\textbf{TriviaQA}\\ \small{July 2017}\end{tabular} &
\begin{tabular}[c]{@{}c@{}}\textbf{BoolQ}\\ \small{May 2019}\end{tabular} &
\begin{tabular}[c]{@{}c@{}}\textbf{NaturalQuestions}\\ \small{July 2019}\end{tabular} &
\begin{tabular}[c]{@{}c@{}}\textbf{TruthfulQA}\\ \small{May 2022}\end{tabular} &
\begin{tabular}[c]{@{}c@{}}\textbf{SelfAware}\\ \small{July 2023}\end{tabular} \\
\midrule
Llama-2-7B-chat-hf        & 29.88\% & 49.78\% & 17.42\% & 20.63\% & 24.64\% \\
Llama-3-8B-Instruct        & 28.69\% & 54.22\% & 17.04\% & 25.00\% & 28.62\% \\
Llama-3.1-8B-Instruct      & 34.66\% & 57.56\% & 22.68\% & 24.38\% & 28.62\% \\
Llama-3.2-3B-Instruct      & 21.91\% & 49.56\% & 17.92\% & 18.75\% & 22.46\% \\
Ministral-8B-Instruct-2410 & 37.45\% & 59.56\% & 21.30\% & 32.50\% & 28.99\% \\
GPT-4o-mini 2024-07-18     & 51.79\% & 77.78\% & 36.97\% & 51.25\% & 40.22\% \\
% Qwen2.5-1.5B-Instruct    & 22.31\% & 54.22\% & 8.40\%  & 16.88\% & 23.55\% \\
% Qwen2.5-3B-Instruct      & 22.31\% & 50.89\% & 14.41\% & 26.88\% & 28.26\% \\
Qwen2.5-7B-Instruct        & 25.90\% & 56.44\% & 18.30\% & 37.50\% & 28.99\% \\
Qwen2.5-14B-Instruct       & 32.67\% & 63.78\% & 24.19\% & 40.63\% & 38.41\% \\
\bottomrule
\end{tabular}}
\caption{Temporal Accuracy (\%): the proportion of time-sensitive questions answered correctly with respect to present-day information, reported across benchmarks.}
\label{tab:temporal-accuracy}
\end{table*}

\begin{table*}[htbp]
\centering
\scalebox{0.9}{
\begin{tabular}{lccccc}
\toprule
\textbf{Model / Dataset (Release Time)} &
\begin{tabular}[c]{@{}c@{}}\textbf{TriviaQA}\\ \small{July 2017}\end{tabular} &
\begin{tabular}[c]{@{}c@{}}\textbf{BoolQ}\\ \small{May 2019}\end{tabular} &
\begin{tabular}[c]{@{}c@{}}\textbf{NaturalQuestions}\\ \small{July 2019}\end{tabular} &
\begin{tabular}[c]{@{}c@{}}\textbf{TruthfulQA}\\ \small{May 2022}\end{tabular} &
\begin{tabular}[c]{@{}c@{}}\textbf{SelfAware}\\ \small{July 2023}\end{tabular} \\
\midrule
Llama-2-7B-chat-hf        & 25.10\% & 57.56\% & 20.18\% & 16.88\% & 19.57\% \\
Llama-3-8B-Instruct        & 28.69\% & 66.00\% & 21.05\% & 32.50\% & 14.86\% \\
Llama-3.1-8B-Instruct      & 38.25\% & 70.00\% & 22.56\% & 26.88\% & 22.10\% \\
Llama-3.2-3B-Instruct      & 24.30\% & 59.78\% & 19.80\% & 17.50\% & 21.01\% \\
Ministral-8B-Instruct-2410 & 29.08\% & 54.67\% & 17.79\% & 31.25\% & 23.19\% \\
GPT-4o-mini 2024-07-18     & 44.22\% & 73.33\% & 23.93\% & 35.63\% & 25.36\% \\
% Qwen2.5-1.5B-Instruct    & 16.73\% & 61.78\% & 8.27\%  & 17.50\% & 14.49\% \\
% Qwen2.5-3B-Instruct      & 17.93\% & 50.89\% & 9.90\%  & 22.50\% & 14.13\% \\
Qwen2.5-7B-Instruct        & 23.11\% & 53.78\% & 15.29\% & 31.25\% & 18.12\% \\
Qwen2.5-14B-Instruct       & 28.69\% & 59.11\% & 17.92\% & 40.00\% & 22.46\% \\
\bottomrule
\end{tabular}}
\caption{Benchmark Fidelity (\%) showing model alignment with benchmark gold answers.}
\label{tab:benchmark-fidelity}
\end{table*}

\subsection{BoolQ $TAG$ Comparison by Controlling Context}
\begin{figure}[H]
    \centering
    \includegraphics[width=0.7\linewidth]{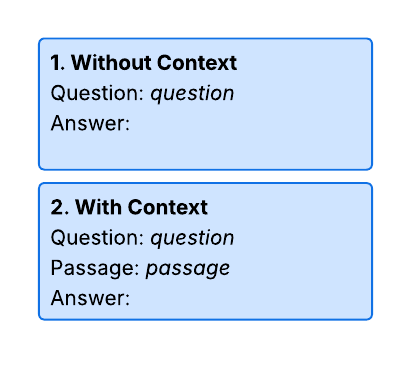}
    \caption{Two prompt formats used in BoolQ experiments: with and without passage context.}
    \label{boolq_context}
\end{figure}

To investigate how outdated context in benchmarks can override updated internal knowledge in LLMs, we conduct controlled experiments on the BoolQ using two prompts, as illustrated in Figure~\ref{boolq_context}. One setting provides both the passage and the question, while the other includes only the question without any supporting passage.

Interestingly, we observe a significant increase in Temporal Alignment Gap when the passage is included. This suggests that although models may have internally updated knowledge, the inclusion of outdated passages often causes them to regress toward older information. Quantitatively, this effect is most pronounced in Ministral-8B-Instruct-2410, which shows a $TAG$ increase of $20.67$ when conditioned on the passage. Similarly, GPT-4o-mini-2024-07-18 exhibits an increase of $19.33$. These large deltas indicate that the models’ updated knowledge is not robust against temporally stale input.

While BoolQ is originally constructed for reading comprehension, our analysis reveals that its static passages can contain outdated facts that actively mislead the model. The $TAG$ between the two settings quantifies the vulnerability of LLMs to temporal anchoring by context, and highlights the need for temporal-awareness in prompt construction and model alignment.

\begin{table}[ht]
\centering
\small
\scalebox{0.9}{
\begin{tabular}{lcc}
\toprule
\textbf{Model} & \textbf{w/o Passage} & \textbf{w/ Passage} \\
\midrule
LLaMA-2-7B-Instruct        & -7.78   & -7.56   \\
LLaMA-3-8B-Instruct       & -11.78  & -16.22  \\
LLaMA-3.1-8B-Instruct      & -12.44  & -21.33  \\
LLaMA-3.2-3B-Instruct      & -10.22  & -16.44  \\
Ministral-8B-Instruct-2410     & \colorbox{Mycolor-green}{4.89 }   & -15.78  \\
GPT-4o-mini-2024-07-18      & \colorbox{Mycolor-green}{4.44 }   & -14.89  \\
% Qwen2.5-1.5B-Instruct      & -7.56   & -18.89  \\
% Qwen2.5-3B-Instruct        & \colorbox{Mycolor-green}{0.00 }   & -14.00  \\
Qwen2.5-7B-Instruct       & \colorbox{Mycolor-green}{2.67 }   & -12.22  \\
Qwen2.5-14B-Instruct       & \colorbox{Mycolor-green}{4.67 }   & -13.56  \\

\bottomrule
\end{tabular}}
\caption{$TAG$ (\%) on BoolQ with and without passage contexts. Positive values (highlighted) indicate alignment between model outputs and current web results, diverging from outdated benchmark gold answers.}
\label{tab:boolq-drift}
\end{table}

\subsection{Model Analysis}

We categorize LLMs into two different groups based on isolated factors to analyze their impacts on temporal misalignment, as shown in Appendix~\ref{appendix:experiment_groups}. 
% We find that a model’s family impacts its internal knowledge, while larger models tend to be more resilient to temporal knowledge drift.
To quantify the impact of time-sensitive questions, we define $TAG$-adjusted accuracy ${a}_{TAG}$ and the $EMR$-adjusted accuracy ${a}_{EMR}$. 
\begin{align}
{a}_{TAG} &= a_o + TAG \cdot \frac{|\mathcal{D}_{\text{ts}}|}{|\mathcal{D}|} \\
{a}_{EMR} &= a_o + EMR \cdot \frac{|\mathcal{D}_{\text{ts}}|}{|\mathcal{D}|}
\end{align}
$a_o$ denotes the LLM accuracy on $\mathcal{D}$.
These adjusted accuracies measure the overall impact of the temporal change of facts in the benchmarks.

\begin{figure}[htbp]
    \centering
    \includegraphics[width=0.88\linewidth]{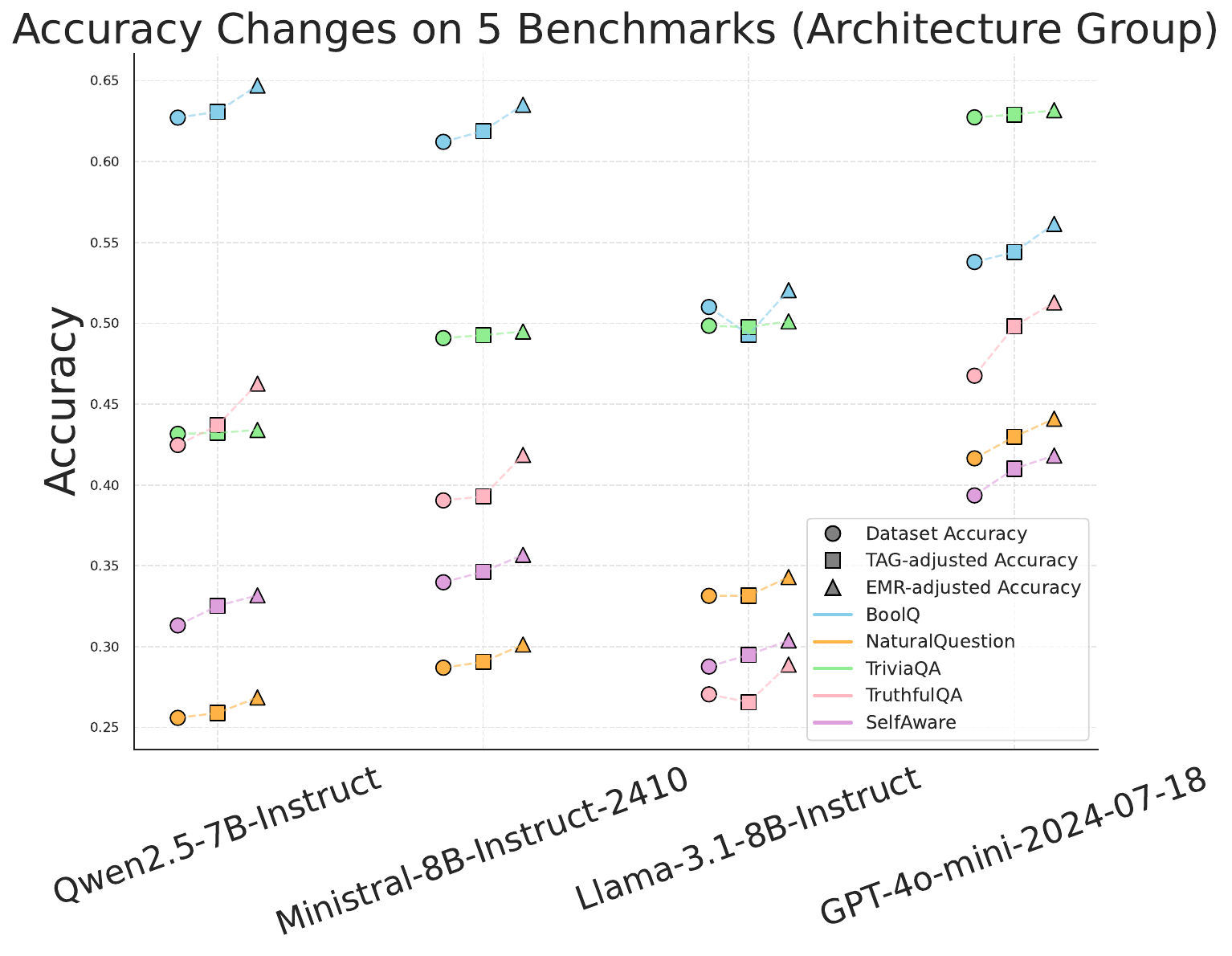}
    \caption{Performance comparison across model architecture families. 
    Accuracy is reported in three settings: Dataset Accuracy, $TAG$-adjusted Accuracy, and $EMR$-adjusted Accuracy.}
    \label{fig:same-family}
\end{figure}

\paragraph{Model Family: The times of memorized facts vary between different LLM families.} 
% topic sentence italic to emphasize
Despite similar release periods and size, LLMs vary in knowledge recency and accuracy, as shown in Figure~\ref{fig:same-family}. 
For example, GPT-4o-mini-2024-07-18 shows a larger performance improvement on the searched information while Llama-3.1-8B-Instruct relies more on outdated answers, indicating that different LLM architectures and training data lead to different times of LLM-memorized facts.
% \begin{figure}[H]
%     \centering
%     \includegraphics[width=0.8\linewidth]{plots/model_group1_comparison_5datasets_dashed.pdf}
%     \caption{Different Model Architecture Families}
%     \label{fig:same-family}
% \end{figure}

% \paragraph{Model Release Time: newer models don't score higher on benchmarks}
% Using the Llama model family, as shown in Table~\ref{tab:main-table},we observed that newer releases don’t necessarily score higher on benchmarks. This is likely due to outdated evaluation data, where models with more current knowledge may be penalized for diverging from obsolete answers. Thus, benchmark scores do not reliably correlate with release dates.

\begin{figure}[htbp]
    \centering
    \includegraphics[width=0.85\linewidth]{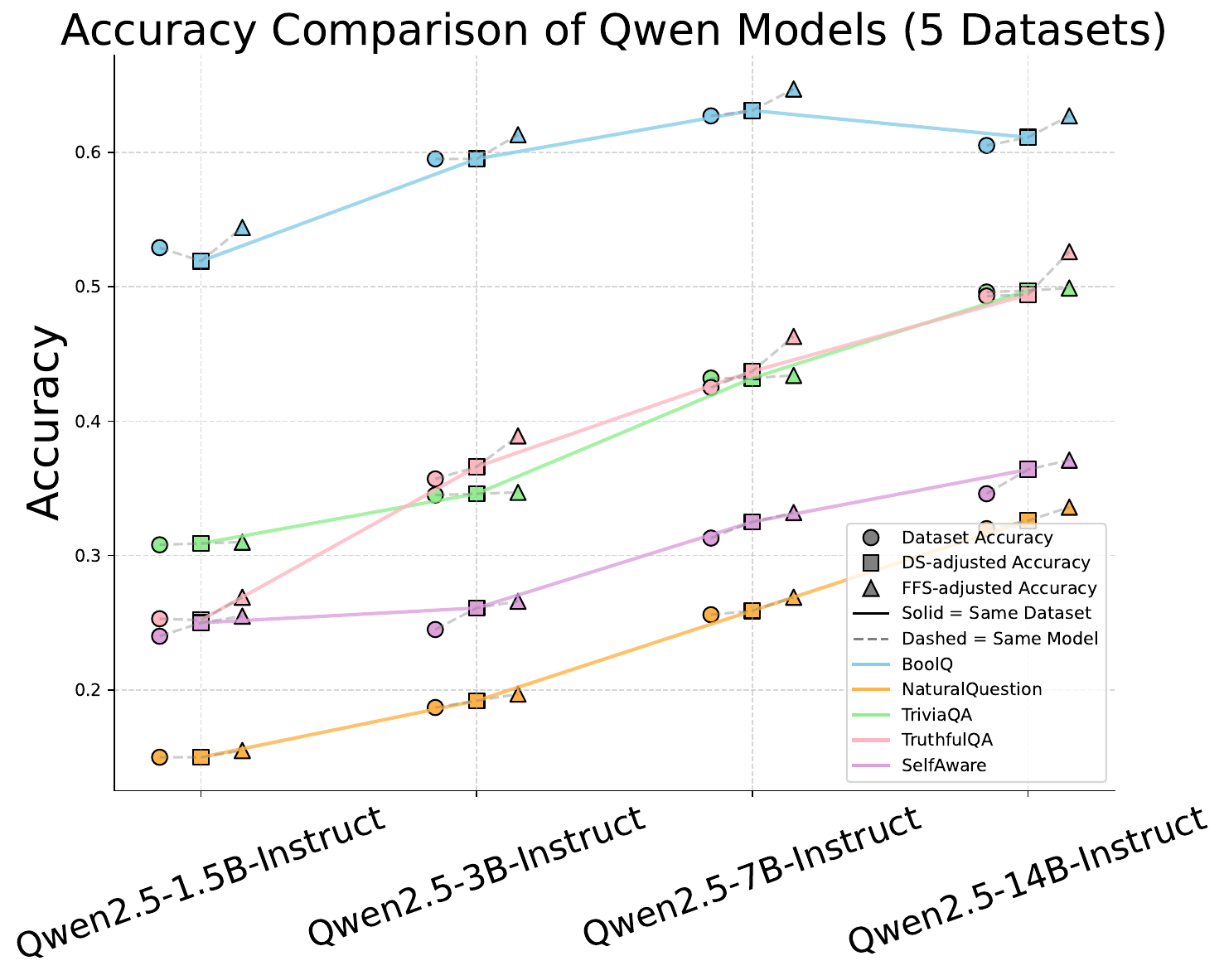}
    \caption{Performance comparison across model sizes within the Qwen family. 
    Accuracy is reported in three settings: Dataset Accuracy, $TAG$-adjusted Accuracy, and $EMR$-adjusted Accuracy.}
    \label{fig5:qwen}
\end{figure}

\paragraph{Model Size: larger models are more robust to time change.}
A study of the Qwen models with different sizes (Figure~\ref{fig5:qwen}) reveals that as model size increases, LLm responses align more with up-to-date searched answers instead of the outdated benchmark answers, suggesting that larger models are more robust and better at adapting to time changes.
We conjecture that more training data for larger models \citep{qwen2025qwen25technicalreport} will cover more recent information.

% \begin{figure*}[htbp]
%     \centering
%     \begin{subfigure}[t]{0.48\linewidth}
%         \centering
%         \includegraphics[width=0.88\linewidth]{plots/model_group1_comparison_5datasets_dashed.pdf}
%         \caption{Different Model Architecture Families}
%         \label{fig:same-family}
%     \end{subfigure}%
%     \hfill
%     \begin{subfigure}[t]{0.48\linewidth}
%         \centering
%         \includegraphics[width=0.85\linewidth]{plots/model_group1_comparison_5datasets_dashed.pdf}
%         \caption{Different Model Sizes (Qwen Family)}
%         \label{fig5:qwen}
%     \end{subfigure}

%     \caption{
%     Performance comparison of LLMs across (a) architecture families and (b) model sizes. 
%     Accuracy is reported in three settings: Dataset Accuracy, $TAG$-adjusted Accuracy, and $EMR$-adjusted Accuracy.
% }
%     \label{fig:combined}
% \end{figure*}

\end{document}